\documentclass[11pt]{article}

\usepackage[preprint]{acl}

\usepackage{times}
\usepackage{latexsym}
\usepackage[T1]{fontenc}
\usepackage[utf8]{inputenc}
\usepackage{microtype}
\usepackage{inconsolata}
\usepackage{graphicx}
\graphicspath{{figures/}}
\usepackage{booktabs}
\usepackage{amssymb}
\usepackage{amsmath}
\usepackage{pifont}
\usepackage[most]{tcolorbox}
\usepackage[table,xcdraw,dvipsnames]{xcolor}
\usepackage{fontawesome5}
\usepackage{enumitem}

\usepackage{listings}

\title{LensLapse: In-Browser Logit Lens for Interactive Exploration \\ of Language Model Training Dynamics}

\author{Tatsuki Okada \quad Yusuke Sakai \quad Taro Watanabe \\
  Nara Institute of Science and Technology (NAIST), Japan \\
  \texttt{okada.tatsuki.ou0@naist.ac.jp, \{sakai.yusuke.sr9, taro\}@is.naist.jp}}

\begin{document}
\maketitle

\begin{abstract}

The logit lens is a standard mechanistic interpretability tool for inspecting a Transformer's predictions across layers. However, existing logit lens tools typically focus on individual checkpoints rather than interactive exploration of a model's \emph{training history}.
In this study, we introduce \textbf{LensLapse}, an open-source web application for interactive exploration of a model's training history through token-level interpretability. Users can scrub a slider across public checkpoints from five model families and three multilingual checkpoint suites and observe how per-layer, per-position logit lens predictions evolve from frequency bias to factual knowledge.
For browser-compatible models, arbitrary user prompts are probed \emph{live in the browser} via local inference over per-checkpoint model pairs. No server, no API key, no hosting cost, and prompts never leave the user's device.
Moreover, \textbf{LensLapse} supports local models. A single \texttt{pip install} enables users to launch the application and explore their own models and training checkpoints without any additional web setup.
LensLapse is MIT-licensed and permanently hosted at \url{https://iamtatsuki05.github.io/lenslapse/}.\footnote{\faGithub \ \ : \href{https://github.com/iamtatsuki05/lenslapse}{\texttt{iamtatsuki05/lenslapse}}}${}^{,}$\footnote{\faPython \ \ : \href{https://pypi.org/project/lenslapse/}{\texttt{pip install lenslapse}}}${}^{,}$\footnote{\faBook \ \ : \href{https://iamtatsuki05.github.io/lenslapse/docs/}{\texttt{iamtatsuki05.github.io/lenslapse/docs}}}${}^{,}$\footnote{\label{footnote:youtube}\faYoutube \ : \url{https://youtu.be/OBnvG9Ru3j8}}
\end{abstract}

\begin{figure*}[t]
  \centering
  \includegraphics[width=0.98\textwidth]{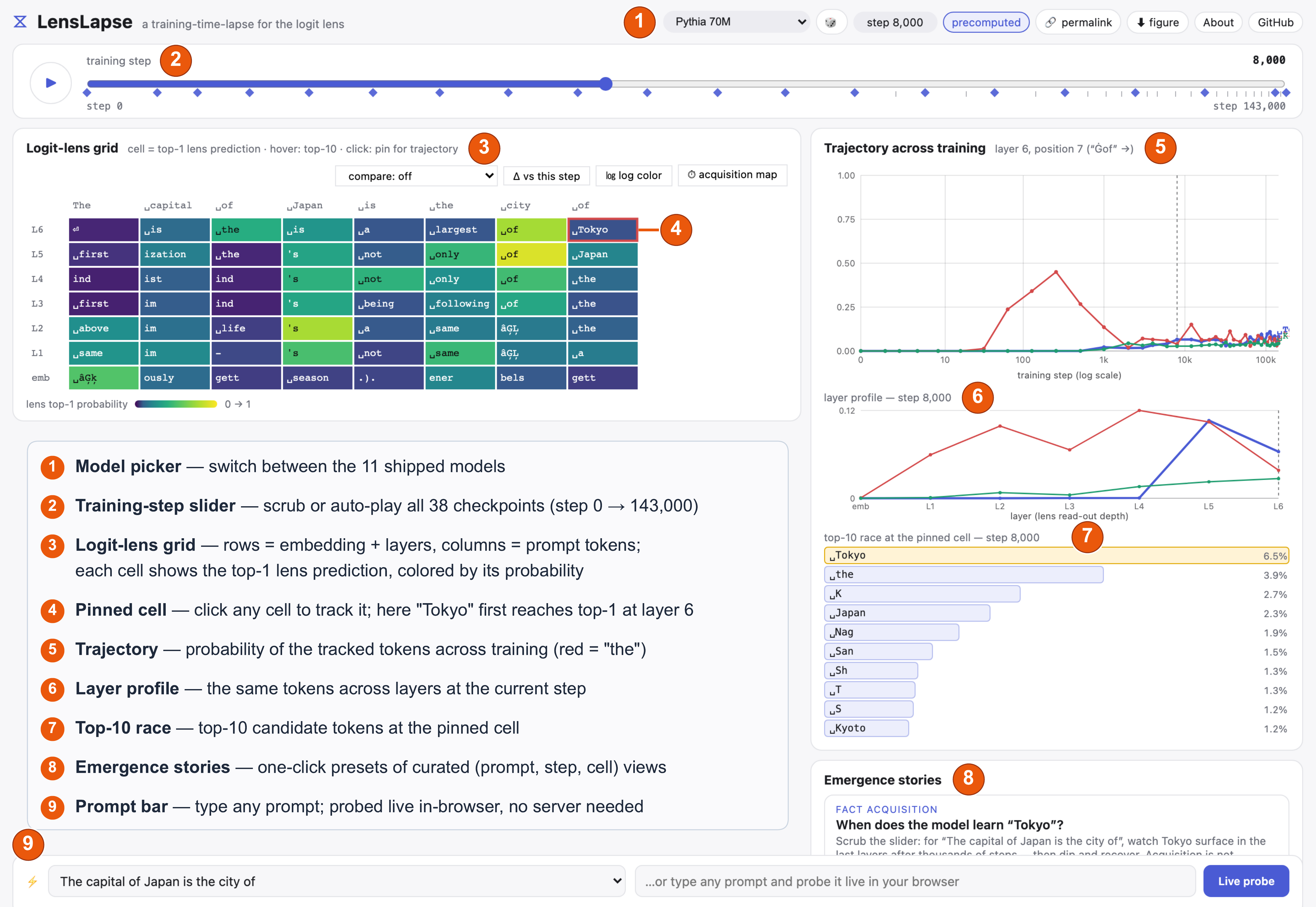}
  \caption{The LensLapse interface at training step 8{,}000 of Pythia-70M for the prompt \emph{``The capital of Japan is the city of''}.
  Components: (1)~model picker over the eleven shipped models; (2)~training-step slider scrubbing the 38 checkpoints (step 0 $\rightarrow$ 143{,}000); (3)~logit lens grid (rows = embedding + 6 layers, columns = token positions; each cell = top-1 lens prediction colored by probability); (4)~pinned cell, where ``\,Tokyo'' first reaches top-1 at layer 6; (5)~trajectory panel plotting exact probabilities across all 38 checkpoints, where the red curve is the frequency-bias token ``\,the'' spiking mid-training and decaying (dashed rule = current step); (6)~layer profile at this step; (7)~top-10 race at the pinned cell; (8)~``emergence stories'' preset views; and (9)~prompt bar probing free text live in-browser.}
  \label{fig:ui}
\end{figure*}

\section{Introduction}

Interpretability tools, e.g., the logit lens \citep{nostalgebraist2020logitlens}, almost exclusively analyze \emph{trained} models, even though the phenomena they reveal, such as factual recall, frequency bias, and layer specialization, emerge \emph{during training}. Checkpoint suites were introduced to facilitate this type of investigation: Pythia ships 154 intermediate checkpoints for each of 16 models trained on identically ordered data~\cite{pmlr-v202-biderman23a}, and OLMo~\cite{groeneveld-etal-2024-olmo} and LLM360~\cite{liu2023llm360fullytransparentopensource} follow the same philosophy. However, in practice, publicly available tools for these checkpoint suites are typically limited to scalar loss curves and static output galleries, while inspecting token-level internal dynamics still requires custom analysis code.
At the same time, interactive interpretability tools are particularly valuable for live demonstrations, exploratory analysis, and communicating model behavior. Hosted lens tools use server-side inference \citep{belrose2025elicitinglatentpredictionstransformers,geva-etal-2022-lm,tufanov-etal-2024-lm}, making their availability dependent on backend maintenance. Client-side execution avoids this requirement for browser-compatible models.

In this study, we introduce \textbf{LensLapse}, an open-source web application for interactive exploration of a model's training history through token-level interpretability.
LensLapse addresses both challenges by combining browser-based inference for browser-compatible models with interactive exploration across training checkpoints. Figure~\ref{fig:ui} shows the interface: a log-scale slider navigates training checkpoints, a grid displays the top-1 lens prediction for every layer-position pair, and clicking a cell visualizes its probability trajectory throughout training. Curated prompts are rendered instantly from precomputed shards, while arbitrary user prompts are processed live on the user’s own hardware. Moreover, it can be easily deployed on user-trained models, facilitating interactive analysis of model behavior throughout training.

The main features of \textbf{LensLapse} are as follows:
\begin{itemize}[leftmargin=*, itemsep=0em, topsep=0.5em, left=0.15em]
  \item  An interactive hosted system for exploring training checkpoints through token-level interpretability over a real public checkpoint suite (three Pythia sizes; 38 precomputed and 20 live-probeable checkpoints each for 14M/70M).
  \item A fully client-side logit lens implementation, exact by construction because the lens over the last block's output reproduces the model's logits, asserted per checkpoint at export; the shipped fp16 format preserves 100\% top-1 agreement with fp32 PyTorch whereas int8 does not (Section~\ref{sec:fidelity}).
  \item A zero-backend architecture based on a static site, versioned weights on the Hugging Face Hub, and client-side computation, enabling zero hosting cost, no inference-server concurrency limit, no server maintenance, and on-device prompt processing.
  \item An open, architecture-agnostic conversion recipe (five decoder-norm families all pass its parity check) that can be applied to arbitrary public checkpoint suites, demonstrated on seven additional public models alongside Pythia and GPT-2: three multilingual suites (MAP-Neo, BAAI Aquila, BLOOM) and four architecture-coverage examples (SmolLM2, Qwen3, OPT, Gemma 3).
\end{itemize}
\begin{table*}[!t]
  \centering
  \small
  \begin{tabular}{lccc}
    \toprule
    System & Lens-style layer view & Training-time axis & Client-side inference \\
    \midrule
    Tuned Lens Space \citep{belrose2025elicitinglatentpredictionstransformers} & \textcolor{Green}{\ding{51}} & \textcolor{Red}{\ding{55}} & \textcolor{Red}{\ding{55}} \\
    LM-Debugger \citep{geva-etal-2022-lm} & \textcolor{Green}{\ding{51}} & \textcolor{Red}{\ding{55}} & \textcolor{Red}{\ding{55}} \\
    LM Transparency Tool \citep{tufanov-etal-2024-lm} & \textcolor{Green}{\ding{51}} & \textcolor{Red}{\ding{55}} & \textcolor{Red}{\ding{55}} \\
    Neuronpedia \citep{neuronpedia} & SAE features & \textcolor{Red}{\ding{55}} & \textcolor{Red}{\ding{55}} \\
    Transformer Explainer \citep{cho2024transformerexplainerinteractivelearning} & \textcolor{Red}{\ding{55}} & \textcolor{Red}{\ding{55}} & \textcolor{Green}{\ding{51}} \\
    BuildAnLLM \citep{buildanllm} & \textcolor{Green}{\ding{51}} & \textcolor{Green}{\ding{51}}\textsuperscript{a} & \textcolor{Red}{\ding{55}} \\
    LLM360 K2 gallery \citep{liu2023llm360fullytransparentopensource} & \textcolor{Red}{\ding{55}} & \textcolor{Green}{\ding{51}} & \textcolor{Red}{\ding{55}} \\
    OLMoTrace \citep{liu-etal-2025-olmotrace} & \textcolor{Red}{\ding{55}} & \textcolor{Red}{\ding{55}}\textsuperscript{b} & \textcolor{Red}{\ding{55}} \\
    \textbf{LensLapse (ours)} & \textcolor{Green}{\ding{51}} & \textcolor{Green}{\ding{51}} & \textcolor{Green}{\ding{51}} \\
    \bottomrule
  \end{tabular}
  \caption{Comparison with interactive interpretability systems. Note that this feature comparison is not a runtime or usability benchmark. Neuronpedia provides SAE feature and circuit views.
  \textsuperscript{a}Supports only local checkpoints.
  \textsuperscript{b}Traces outputs to training \emph{data}, rather than across training \emph{time}.}
  \label{tab:related}
\end{table*}
LensLapse targets interpretability and training-dynamics researchers browsing hypotheses before writing analysis code, instructors and students without GPU access, and checkpoint-suite maintainers seeking a reusable front-end for hosting their own models and building interactive demonstrations.

\section{Background and Related Work}
\label{sec:related}

Table~\ref{tab:related} compares LensLapse with similar interactive interpretability systems. Hosted logit lens tools inspect a single trained model using server-side inference \citep{belrose2025elicitinglatentpredictionstransformers,geva-etal-2022-lm,tufanov-etal-2024-lm}, while Neuronpedia~\citep{neuronpedia} provides a rich hosted platform for Sparse Autoencoder (SAE) features and circuits for trained models. Transformer Explainer \citep{cho2024transformerexplainerinteractivelearning} introduced in-browser GPT-2 inference for education but does not expose layer-wise vocabulary predictions. BuildAnLLM \citep{buildanllm} provides a local checkpoint viewer for user-trained models. OLMoTrace \citep{liu-etal-2025-olmotrace} traces outputs back to training \emph{data}, complementing our focus on training \emph{time}. Recent work tracks representation formation throughout pretraining using crosscoders \citep{bayazit2026crosscodingtimetrackingemergence,ge2026evolutionconceptslanguagemodel}. LensLapse complements these methods by providing browser-based logit-lens visualization across checkpoints, without implementing crosscoder analysis.

\section{LensLapse: System Description}
\label{sec:system}

\subsection{Interface}

As shown in Figure~\ref{fig:ui}, LensLapse provides a single-page interface organized with three coupled views: %

\begin{description}[leftmargin=*, itemsep=0em, topsep=0.5em, left=0em]
\item[Lens grid] renders the top-1 lens prediction of the selected checkpoint for every (layer, position) cell on a canvas, colored by probability; hovering reveals top-10 distribution, and clicking pins a cell. 
\item[Training-step slider] is log-scaled with one tick for each precomputed checkpoint; diamonds indicate live-probeable checkpoints. Scrubbing the slider, or using one-click playback, animates the grid, with cells flashing as their top-1 predictions change. An \emph{acquisition-map} view recolors the grid according to when each cell first predicts its final answer, while a \emph{diff} view highlights changes relative to a fixed reference checkpoint.
\item[Trajectory panel] plots, for the pinned cell, the exact probabilities across every checkpoint for the tokens to which the final model converges (the final top-3 plus the gold continuation; typed tokens can be added to the live-traced set). It also displays the current checkpoint as a movable rule, together with an animated top-10 bar race and the classic p-vs-layer profile.
Curated \emph{emergence stories} jump to preset (prompt, checkpoint, pinned-cell) configurations for first-time users. The full application state is encoded in the URL hash, making every view a shareable permalink, while one-click export saves the current view as a publication-ready PNG or PDF figure.
A model picker switches among all eleven shipped models, each labeled with its documented language support. Grid depth, checkpoint list, and live-probeable checkpoints adapt automatically, and a compare mode renders a second model's grid in lockstep for side-by-side comparison across model scales.
\end{description}

\subsection{Efficient Hosting and Data Flow}
We support two complementary execution modes. For common use cases, we provide precomputed prompts and hosted models optimized for efficient browser-based execution. For larger models or user-trained checkpoints, we also provide an easy deployment pipeline, enabling the same interactive interface through a lightweight local backend.

\paragraph{Precomputed path.}
For 16 curated prompts, or 26 for the multilingual-capable models, we precompute the full lens output using fp32 PyTorch over 38 checkpoints (20 for the 160M model) and serve the results as static JSON shards. Each shard contains the top-10 tokens for every cell, together with exact probability and rank trajectories for position-wise target tokens. This enables instant visualization without downloading model weights; the grid renders in under one second on the first visit.

\paragraph{Live path.}
Any free-text prompt triggers a forward pass, without autoregressive generation or KV caching, through every registered checkpoint rather than only the selected one, allowing the same time-lapse playback available for curated prompts.
Model files are fetched once from a public Hugging Face repository, cached via the browser Cache API, and executed with ONNX Runtime Web \citep{onnxruntime}, preferring WebGPU and falling back to WASM or, if model files are unavailable, to a precomputed-only mode with a visible status badge. Tokenization uses Transformers.js \citep{transformersjs}.
All (layer, position) hidden states are passed through the lens head using a single batched matrix multiplication per checkpoint, while softmax and top-10 extraction are performed in JavaScript.
Because inference runs entirely on the client, user prompts never leave the device. We use ONNX and JavaScript rather than PyTorch-based TransformerLens, nnsight, or tuned-lens to avoid requiring a Python backend in this path.

\paragraph{Heavy models and reproducible probes.}
We benchmark browser inference only for Pythia-14M--160M. We currently provide off-the-shelf support for models up to Qwen3-0.6B. The maximum browser-compatible size and the first limiting factor (download size, memory, or latency) are not considered. When browser resources are insufficient, precomputed views or a probe server are needed. MAP-Neo and Aquila also need the server for free-text tokenization.
For local use, a single \texttt{pip install lenslapse} provides the application and probe API through a single local port, without CORS configuration, and supports offline use once models are cached.
The interface registers Hub models and local runs persistently without export, supports background ONNX conversion, and detects a local server when available.
The server uses Hugging Face Transformers \citep{wolf-etal-2020-transformers} and the precomputed pipeline's hooked forward pass to obtain all layers' residual-stream states. %
A \emph{trace-across-training} action probes every registered checkpoint for trajectory visualization. UI actions are scriptable through the \texttt{lenslapse} CLI (\texttt{probe}, \texttt{trace}, and \texttt{models}).
Every live result is cached in IndexedDB (browser) and a keyed JSON cache (server), enabling byte-for-byte replay across sessions. Appendix~\ref{app:integration} provides usage examples.

\subsection{Zero-Backend Deployment}

The site is a static bundle on GitHub Pages\footnote{The site is up to $\sim$250MB, with tokenizer files for the multilingual models and precomputed shards accounting for most of the size, well below the GitHub Pages 1GB limit.} and model weights live in a public Hugging Face repository and are fetched directly by the browser.
Nothing executes server-side, so hosting is free, capacity scales with the number of visitors' devices, and there is no backend to maintain. Therefore, once deployed, the application can remain available without server maintenance.

\paragraph{Bring your own suite.}
Adding a model requires only data preparation rather than code modification. A single onboarding command (\texttt{lenslapse add-model}) exports, parity-checks, precomputes, and registers a Hugging Face checkpoint suite, a single Hub model, or a local directory of Trainer checkpoints, including user-trained models. The resulting ONNX models can then be deployed on any static hosting service.
The included GitHub Actions workflow enables end-to-end deployment with no hosting cost (see \texttt{docs/deployment.md}). The full pipeline, e.g., conversion, precomputation, fidelity evaluation, and the benchmark harness, is released under MIT license in the same repository.

\paragraph{Blog-style interactive documents.}
Because precomputed results are stored as static JSON files, the same framework can be readily adapted to arbitrary datasets and models. Even for models too large for browser inference, interactive visualizations can be embedded as \textbf{\textit{precomputed path}}. This makes it practical to publish interactive analyses alongside papers, blogs, documentation, or entire datasets, improving accessibility, reproducibility, and long-term availability of research artifacts. 

\section{Evaluation}
\label{sec:eval}

\subsection{Numerical Fidelity}
\label{sec:fidelity}

Table~\ref{tab:fidelity} compares different weight formats against fp32 PyTorch ground truth over all (layer, position) cells for 16 curated prompts at three checkpoints. Dynamic int8 quantization, a common choice for browser deployment, is unsuitable in this setting: final-layer top-1 agreement drops to 52\% (per-tensor) or 71\% (per-channel) at step 143k, and the degradation increases throughout training as weight outliers emerge. Consequently, quantization errors are largest at later checkpoints, where inspection is often most informative.
In contrast, storing weights in fp16 while computing in fp32 preserves 100\% top-1 agreement on this benchmark while reducing storage by half relative to fp32, and is therefore the default configuration in LensLapse.

In the production browser implementation (WebGPU), live probing matches the fp32 PyTorch precomputed results on all 56 evaluated cells, with a maximum absolute probability difference of $5.5{\times}10^{-4}$. The ONNX-free probe server (fp32 by default) also achieves identical top-1 predictions on all 56 cells. By comparison, an fp16 server-side forward pass already flips 5 of the 56 predictions at step 143k, and lower-precision execution is therefore supported only as an explicit opt-in.
In addition, capturing raw block outputs avoids double normalization of the final hidden state. Further details are provided in Appendix~\ref{sec:recipe}. However, numerical fidelity does not establish interpretive reliability: intermediate-layer predictions can be brittle \citep{belrose2025elicitinglatentpredictionstransformers}. Tuned-lens support and cross-family validation of interpretive reliability remain future work.

\begin{table}[t]
  \centering
  \small
  \setlength{\tabcolsep}{3.5pt}
  \begin{tabular}{lcccr}
    \toprule
    & \multicolumn{3}{c}{Top-1 agreement (all / final layer)} & \\
    \cmidrule(lr){2-4}
    Format & step 8k & step 64k & step 143k & MB \\
    \midrule
    fp32 ONNX & 1.00 / 1.00 & 1.00 / 1.00 & 1.00 / 1.00 & 283 \\
    int8 (tensor) & .90 / .88 & .73 / .69 & .61 / \textbf{.52} & 71 \\
    int8 (channel) & .93 / .95 & .78 / .85 & .74 / .71 & 72 \\
    \textbf{fp16 (ours)} & \textbf{1.00 / 1.00} & \textbf{1.00 / 1.00} & \textbf{1.00 / 1.00} & 142 \\
    \bottomrule
  \end{tabular}
  \caption{Lens fidelity vs.\ fp32 PyTorch on Pythia-70M (16 prompts; 861 cells per checkpoint). Mean final-layer KL at step 143k: $2{\times}10^{-5}$ (fp32, fp16), 0.18 (int8 channel), 0.60 (int8 tensor). The fp32 size includes the backbone's external weight file.}
  \label{tab:fidelity}
\end{table}

\begin{figure}[t]
  \centering
  \includegraphics[width=\linewidth]{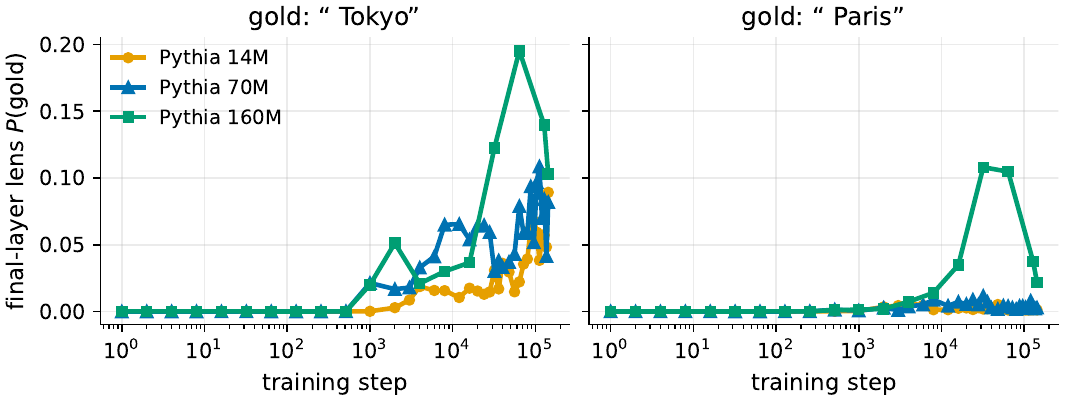}
  \caption{Scale $\times$ time in one view: exact final-layer lens probability of the gold continuation across training for all three shipped Pythia models, read from the precomputed shards (markers = checkpoints, log-scale steps). Left: ``\,Tokyo'' is acquired by every size, non-monotonically. Right: ``\,Paris'' is acquired only by 160M, a capability threshold, and later decays.}
  \label{fig:scale}
\end{figure}

\subsection{Performance across Device Settings}
\label{sec:perf}

Table~\ref{tab:perf} reports latency across browser engines, execution providers, emulated CPU throttling, model sizes, and prompt lengths, measured using our scripted Playwright harness.
A \emph{probe} measures the end-to-end latency from pressing ``Live probe'' to rendering the results, including a single forward pass, lens-head computation over all cells, and JavaScript softmax/top-10 extraction.
Warm probes for 8-token prompts remain interactive ($\leq$320\,ms) on unthrottled Chromium and WebKit across all three model sizes. Firefox using the WASM backend is the only notable outlier ($\sim$1.4\,s, about 11$\times$ slower than Chromium-WASM), reflecting differences in ONNX Runtime's WASM implementation.
WASM latency scales approximately linearly with model size (66/127/316\,ms warm for 14M/70M/160M), whereas WebGPU shows sub-linear scaling (63/70/134\,ms). Under 4$\times$ CPU throttling (DevTools emulation of a low-end laptop), WASM slows by about 4.2$\times$ and WebGPU by about 3.3$\times$. Because top-10 extraction is performed on the CPU, even the WebGPU pipeline retains a CPU-bound component.

\begin{table}[t]
  \centering
  \small
  \setlength{\tabcolsep}{2pt}
  \begin{tabular}{@{}llllrr@{}}
    \toprule
    Browser & EP & Model & Thr. & Load c/w & Probe f/w \\
    \midrule
    Chromium & WebGPU & 70M & 1$\times$ & 1052 / 295 & 157 / \textbf{70} \\
    Chromium & WASM & 70M & 1$\times$ & 935 / 230 & 155 / \textbf{127} \\
    Chromium & WebGPU & 70M & 4$\times$ & 3224 / 1066 & 488 / \textbf{233} \\
    Chromium & WASM & 70M & 4$\times$ & 2821 / 888 & 632 / \textbf{530} \\
    Chromium & WebGPU & 14M & 1$\times$ & 566 / 106 & 149 / \textbf{63} \\
    Chromium & WASM & 14M & 1$\times$ & 509 / 84 & 94 / \textbf{66} \\
    Chromium & WebGPU & 160M & 1$\times$ & 1799 / 590 & 237 / \textbf{134} \\
    Chromium & WASM & 160M & 1$\times$ & 1610 / 489 & 352 / \textbf{316} \\
    Firefox & WASM & 70M & 1$\times$ & 2404 / 2850 & 2320 / \textbf{1399} \\
    WebKit & WASM & 70M & 1$\times$ & 644 / 203 & 160 / \textbf{125} \\
    Chromium & WebGPU & 70M\textsuperscript{\dag} & 1$\times$ & 1045 / 279 & 310 / \textbf{205} \\
    Chromium & WASM & 70M\textsuperscript{\dag} & 1$\times$ & 916 / 231 & 509 / \textbf{453} \\
    \bottomrule
  \end{tabular}
  \caption{Latency in ms on an M1 Max (2026-07-15). Load = checkpoint fetch (local host) + session init, cold / Cache-API-warm. Probe = forward + lens + top-10, first / warm mean of 5. Thr.\ = emulated CPU throttling. \textsuperscript{\dag}with a 30-token prompt (all other rows use 8-token prompts). }
  \label{tab:perf}
\end{table}

\begin{figure*}[t]
  \centering
  \begin{minipage}[t]{0.47\textwidth}
    \centering
    \includegraphics[width=\linewidth]{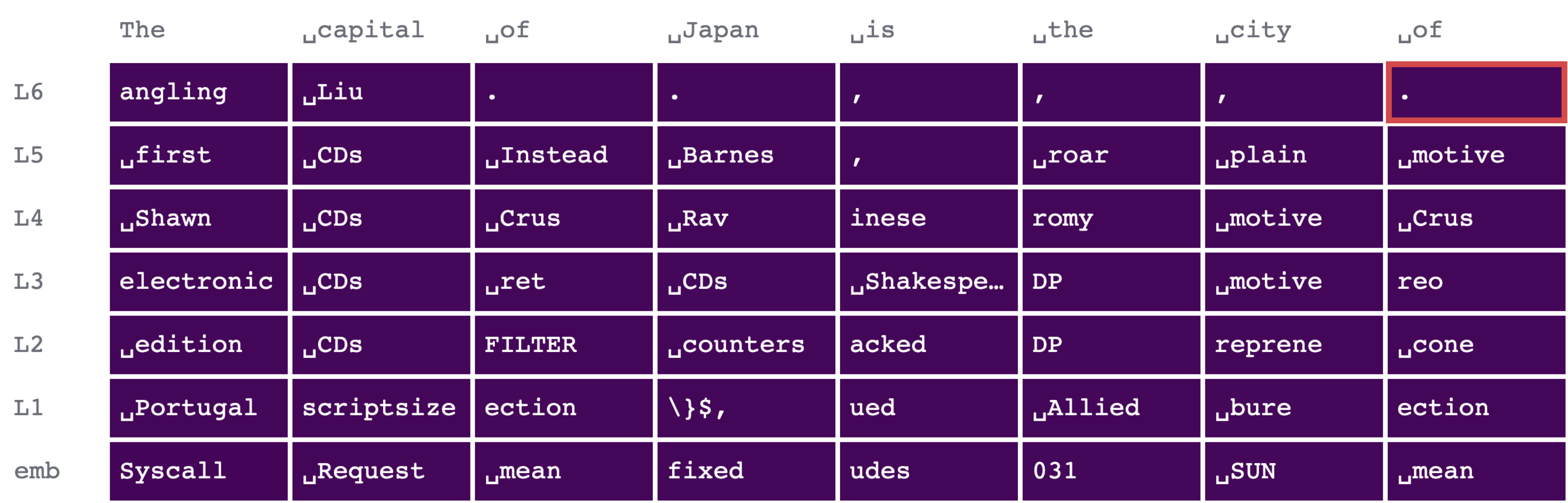}\par\smallskip
    {\small (a) step 16: near-uniform noise (top-1 $p\!\approx\!3{\times}10^{-4}$)}
  \end{minipage}\hfill
  \begin{minipage}[t]{0.47\textwidth}
    \centering
    \includegraphics[width=\linewidth]{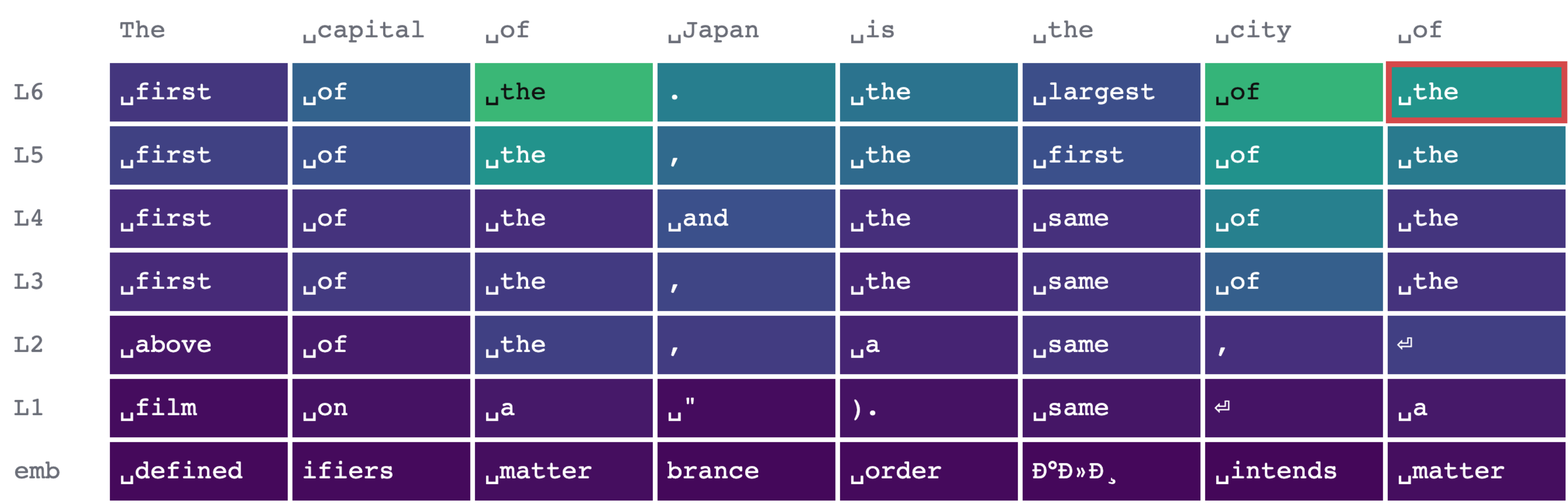}\par\smallskip
    {\small (b) step 512: high-frequency tokens dominate}
  \end{minipage}\par\medskip
  \begin{minipage}[t]{0.47\textwidth}
    \centering
    \includegraphics[width=\linewidth]{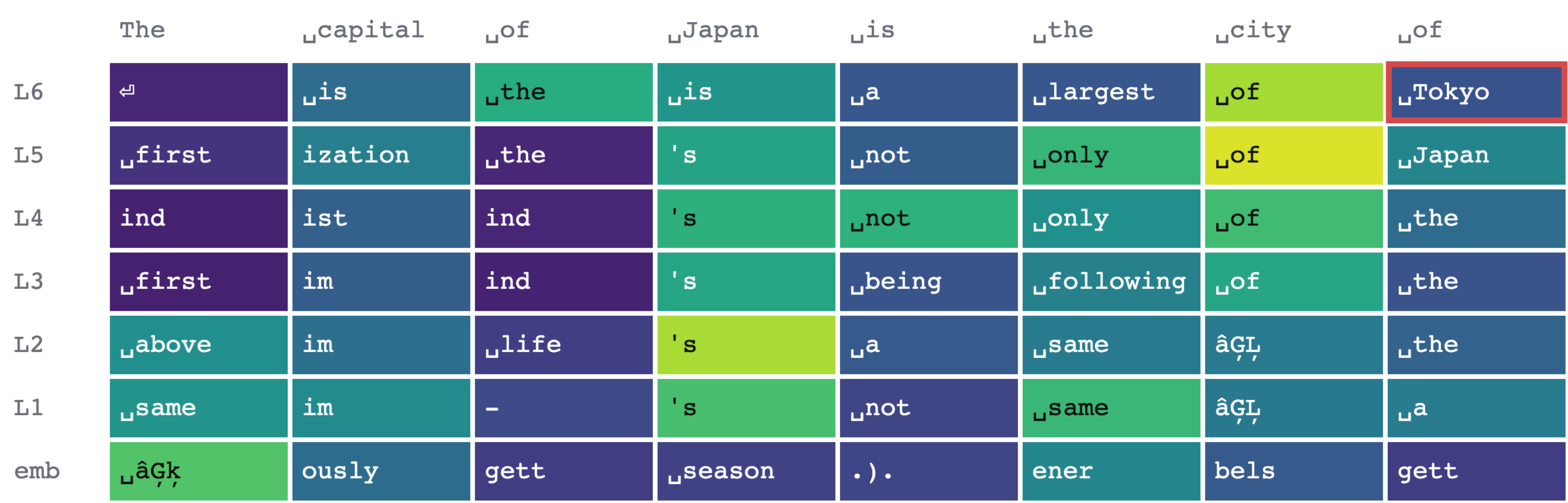}\par\smallskip
    {\small (c) step 8{,}000: ``\,Tokyo'' first reaches top-1 (L6)}
  \end{minipage}\hfill
  \begin{minipage}[t]{0.47\textwidth}
    \centering
    \includegraphics[width=\linewidth]{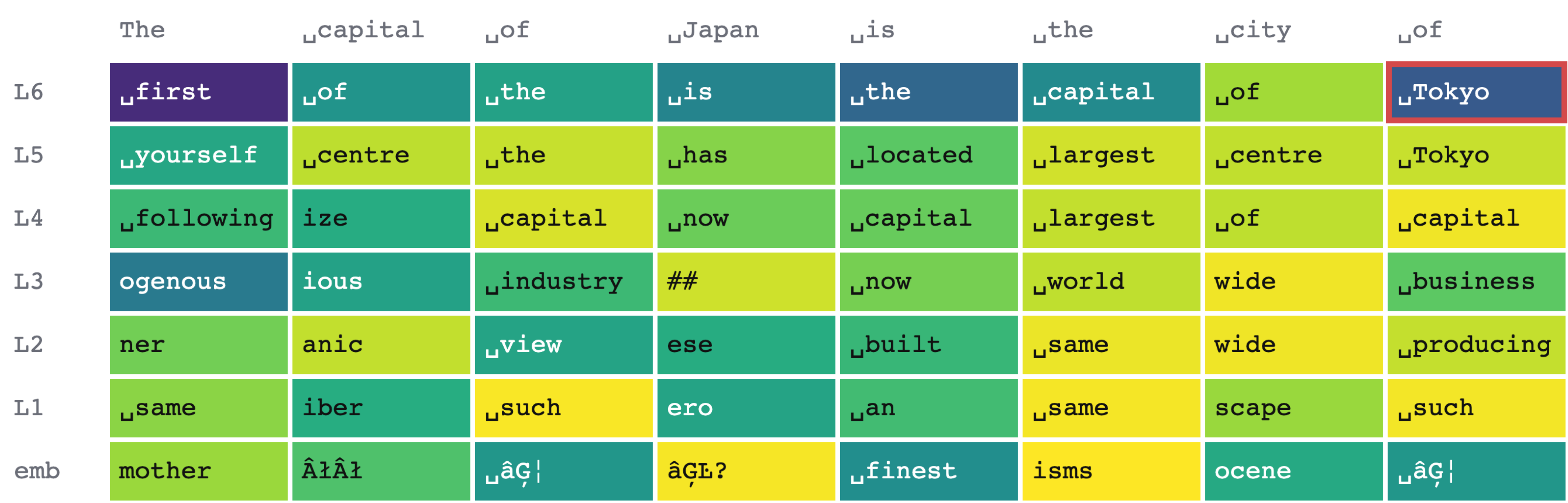}\par\smallskip
    {\small (d) step 143{,}000: facts settled in the deep layers}
  \end{minipage}
  \caption{The time-lapse: the lens grid of Pythia-70M for \emph{``The capital of Japan is the city of''} at four training steps (rows = embedding + layers bottom-up, columns = token positions, color = top-1 probability; the outlined cell is the final position at layer 6). Scrubbing the slider animates this progression continuously over 38 checkpoints.}
  \label{fig:strip}
\end{figure*}

\begin{figure}[t]
  \centering
  \includegraphics[width=1.1\linewidth]{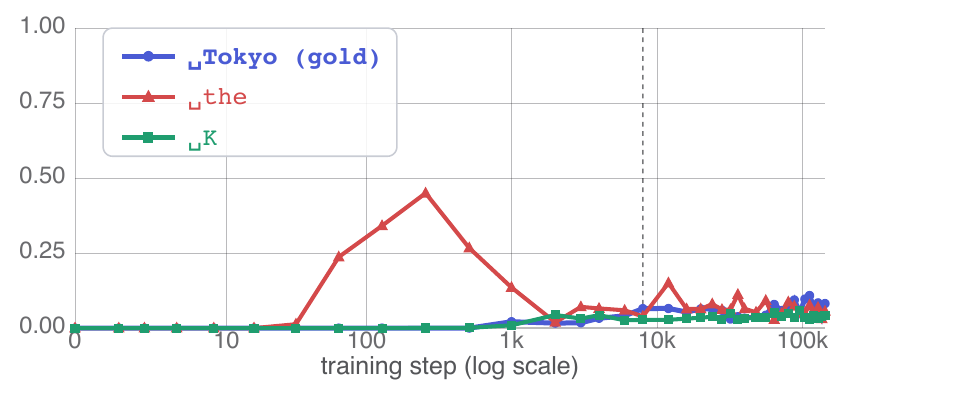}
  \caption{Exact lens-probability trajectories for the pinned cell of Figure~\ref{fig:strip} across all 38 checkpoints: ``\,the'' (red) spikes mid-training and decays, ``\,Tokyo'' rises late and non-monotonically (dashed rule = selected step).}
  \label{fig:traj}
\end{figure}

\subsection{Case Studies}
\label{sec:cases}

Figure~\ref{fig:strip} shows four representative time-lapse snapshots. All observations below are obtained directly from LensLapse and are reproducible via permalinks. Unless otherwise noted, probabilities refer to the final-layer lens at the last prompt position of Pythia-70M.

\paragraph{Fact acquisition is visible and non-monotonic.}
For \emph{``The capital of Japan is the city of''}, $P(\text{``\,Tokyo''})$ remains near zero until approximately step 1k and first reaches top-1 at step 8k. However, the trajectory is far from monotonic: across the 38 checkpoints, Tokyo is top-1 at only 10 checkpoints, losing the position throughout most of steps 12k--56k (bottoming at $P{=}0.030$ at step 32k) before recovering to $P{=}0.082$ at step 143k. In addition, Figure~\ref{fig:traj} shows the trajectory panel. It also shows $P(\text{``\,the''})$ peaking during mid-training before declining, indicating competition between frequency bias and factual knowledge.

\paragraph{Early frequency bias.}
At step 16 the lens distribution is nearly uniform (top-1 $p \approx 3{\times}10^{-4}$). By step 64, the five most frequent predictions (``,'', ``\,the'', ``.'', newline, and ``\,and'') account for 67\% of all 861 grid cells in the 16-prompt suite. By step 512, this top-5 share falls to 36\%, before more position-specific predictions emerge.

\paragraph{Layer specialization emerges gradually.}
At step 143k, ``\,Tokyo'' is top-1 only at layers 5--6, while lower layers predict locally plausible yet incorrect continuations (``\,capital'' at L4 and ``\,business'' at L3). Earlier checkpoints show even the deepest layers have not yet acquired the fact.

\paragraph{Browsing acquisition order.}
Reading first-acquisition steps from the gold continuations reveals a consistent progression: code completion first (\emph{``return a +''} $\rightarrow$ ``\,b'': step 1k), followed by idioms and agreement (``\,much'': 2k, ``\,time'': 3k), and finally factual knowledge (``\,Tokyo'': 8k, ``\,Washington'': 32k). This progression from syntax to semantics to factual knowledge is observable within a single interface.

\paragraph{Learned, then lost.}
``\,oxygen'' (\emph{``Water is made of hydrogen and''}) becomes top-1 at step 8k but loses the position by step 143k. ``\,Washington'' exhibits a similar pattern, illustrating that knowledge acquisition is not monotonic. LensLapse makes such knowledge-drift patterns straightforward to inspect interactively.

\paragraph{Scale and training time in one view.}
Figure~\ref{fig:scale} illustrates how model scale interacts with training dynamics. The more difficult fact \emph{``The Eiffel Tower is located in the city of''} $\rightarrow$ ``\,Paris'' is never acquired as top-1 by the 14M model (final $P{=}0.0012$) or the 70M model ($P{=}0.0028$), whereas the 160M model reaches top-1 at step 32k before falling to rank 3 by step 143k ($P{=}0.022$), exhibiting another learned-then-lost trajectory. In contrast, ``\,Tokyo'' is acquired by all three model sizes, albeit at different and similarly non-monotonic stages of training (14M: 72k, 70M: 8k, 160M: 32k).

\section{Conclusion}

In this study, we introduced \textbf{LensLapse}, an open-source web application for interactive exploration of training dynamics through token-level interpretability. LensLapse provides high-fidelity browser-based visualization of logit lens predictions, supports user-trained models through a lightweight local probe server, and enables sustainable zero-backend deployment. Our experiments demonstrate that LensLapse enables interactive inspection of diverse training phenomena. We hope it will facilitate research, education, and interactive dissemination of language model interpretability.

\appendix

\section{Installation and Deployment}
\label{app:integration}

Listing~\ref{lst:install} shows the complete local setup. A single \texttt{pip install} and one command launch both the web application and the live probe API on the same port, without requiring ONNX conversion or configuration files. Listing~\ref{lst:deploy} shows how to publish a permanent, zero-cost deployment with user-trained models using the same \texttt{add-model} onboarding command described in Section~\ref{sec:system}.

\begin{lstlisting}[language=sh, float=b, caption = {Installing and probing your own model locally: one command serves the app and its probe API on the same port.}, label = lst:install]
pip install lenslapse
lenslapse server
# Serving app + probe API at http://localhost:8017/
# Opening in your default browser...

# In the header: click the gear icon, enter a Hugging Face model id
# (or browse to a local training-checkpoint folder), and probe it live --
# no ONNX export, no config files.
\end{lstlisting}

\begin{lstlisting}[language=sh, float=!t, caption = {Deploying your own permanent, zero-cost copy of the site with your own model added.}, label = lst:deploy]
gh repo fork iamtatsuki05/lenslapse --clone && cd lenslapse

uv run lenslapse add-model --model <your-hf-model-or-local-checkpoint-dir> \
  --id my-model --label "My Model" --models-root ./onnx-out
# -> exports + parity-checks the ONNX pair, precomputes the lens shards,
#    and registers "my-model" in web/public/data/models.json
#    (a data change, not a code change).

hf upload-large-folder <you>/my-model-onnx --repo-type model ./onnx-out
# then point HF_DEFAULT in web/src/live.ts at <you>/my-model-onnx

git push
# One-time only: Settings -> Pages -> Source: GitHub Actions.
# -> deploy-pages.yml builds web/ (now including "my-model") and publishes
#    it live at https://<you>.github.io/lenslapse/ -- no server to run.
\end{lstlisting}

\section{Conversion Recipe}
\label{sec:recipe}

Each checkpoint is exported as a pair of ONNX graphs.
The \textbf{backbone} maps token ids to the residual stream after the embedding and after each transformer block, $[L{+}1, T, H]$.
A subtlety makes correctness non-trivial: Hugging Face Transformers applies the final layer norm to the last entry of \texttt{output\_hidden\_states}, so naively lensing it double-normalizes (we measured a maximum logit error of 0.59).
We instead capture each block's raw output with forward hooks, giving pre-norm states uniformly.
The \textbf{lens head} is a separate three-node graph (final norm + unembedding) applicable to any layer's hidden state; splitting it from the backbone avoids duplicating the unembedding matrix.
This yields an exactness identity, $\mathrm{lens}(h_L) \equiv$ model logits, which the export pipeline asserts per checkpoint: the difference is exactly $0.0$ in PyTorch, and the exported models preserve top-1 at every probe position, with a maximum logit deviation of $1.1{\times}10^{-2}$ across the 60 Pythia fp16 checkpoints; the GPT-2 export passes the same assertions.

Weights are stored as fp16 initializers with a runtime cast that ONNX Runtime constant-folds at session load, halving file size (29/142/326\,MB per checkpoint for 14M/70M/160M) while keeping computation in fp32.
The recipe is architecture-generic: it introspects the decoder stack, the final norm (LayerNorm or RMSNorm), and the unembedding, and a parity checker asserts the lens identity per model.
GPT-NeoX (Pythia, PyTorch diff 0.0), GPT-2-style LayerNorm (GPT-2 \citep{radford2019language} with tied embeddings; BLOOM's ALiBi variant), Llama-style RMSNorm (SmolLM2-135M, 30 layers, \citealp{allal2025smollm2smolgoesbig}; MAP-Neo-250M), Mistral-style RMSNorm (Aquila-135M; Qwen3-0.6B), and Gemma-style plus-one-weight RMSNorm (Gemma-3-270M) all pass; unsupported (e.g., post-norm) layouts fail loudly at export rather than shipping a wrong lens.
A layout the generic attribute-path heuristic cannot reach, OPT's decoder nests two hops deeper than the others, registers an explicit override for that architecture's own \texttt{config.model\_type} in one line, replacing the heuristic entirely for it rather than blending with it, so a wrong registration also fails loudly rather than silently guessing.

\end{document}